\documentclass[letterpaper, 10 pt, conference]{ieeeconf}  

\IEEEoverridecommandlockouts                              

\usepackage{graphics} 
\usepackage{epsfig} 
\usepackage{times} 
\usepackage{amsmath} 
\usepackage{amssymb}  
\usepackage{cite}
\usepackage{subcaption}
\usepackage{hyperref}
\usepackage{cleveref}
\usepackage{booktabs}
\usepackage{xcolor}
\usepackage{pifont}
\usepackage{caption}
\usepackage[table]{xcolor} 
\usepackage{tikz}
\usepackage{multirow} 
\definecolor{lightblue}{RGB}{224,255,255}
\definecolor{lightpink}{RGB}{255, 230, 230}  %
\newcommand{\qheading}[1]{\noindent\mbox{\textbf{#1}\;}}

\title{\LARGE \bf
Bi-Adapt: Few-shot Bimanual Adaptation for Novel Categories of \\ 3D Objects via Semantic Correspondence
}

\author{
\authorblockN{
Jinxian Zhou\textsuperscript{1,2*},
Ruihai Wu\textsuperscript{3*}, 
Yiwei Liu\textsuperscript{1},
Yiwen Hou\textsuperscript{1}, \\
Xunzhe Zhou\textsuperscript{1,4},
Checheng Yu\textsuperscript{1,4},
Licheng Zhong\textsuperscript{2},
Lin Shao\textsuperscript{1$\dagger$}
}
\authorblockA{
\textsuperscript{1}National University of Singapore \quad
\textsuperscript{2}Shanghai Qi Zhi Institute \quad
\textsuperscript{3}Peking University \quad
\textsuperscript{4}The University of Hong Kong
}
}
\begin{document}
\thispagestyle{empty}
\pagestyle{empty}

\twocolumn[{
\renewcommand\twocolumn[1][]{#1}
\maketitle

\begin{center}
    \captionsetup{type=figure}
    \includegraphics[width=0.95\textwidth]{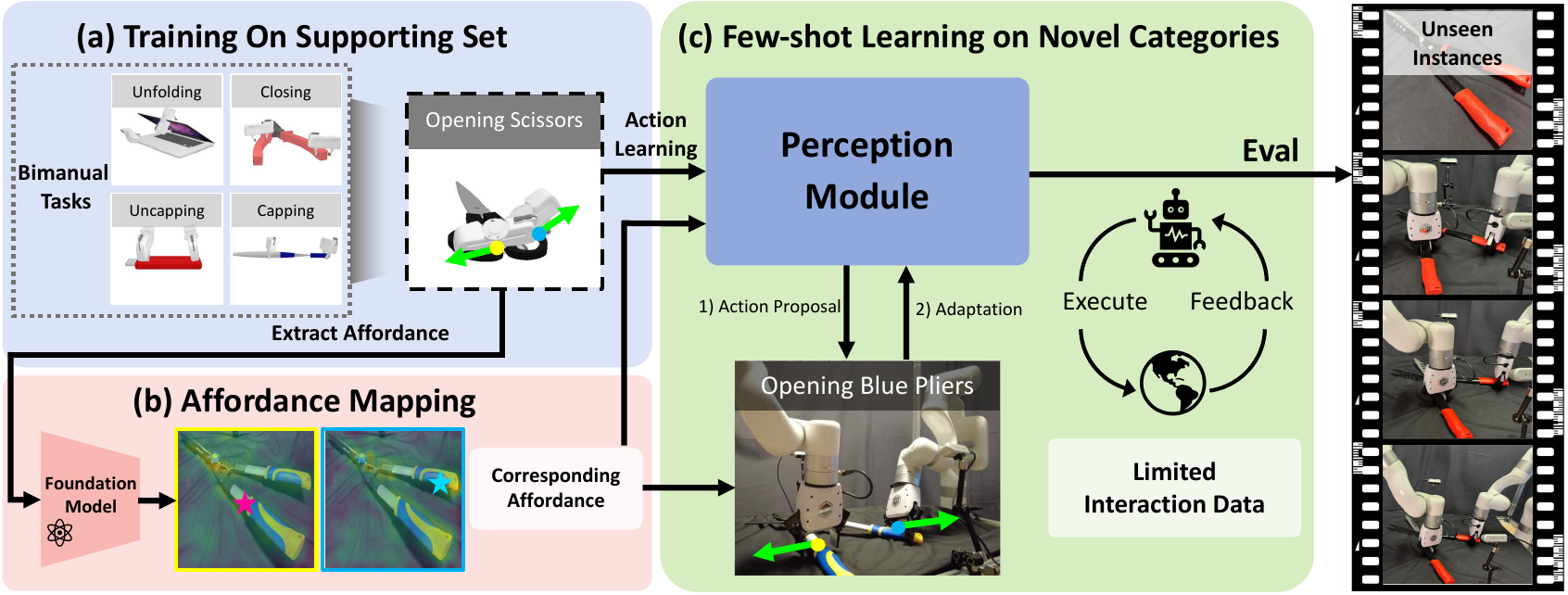}
\caption{
We present \textbf{Bi-Adapt}, a novel framework designed for efficient learning of generalizable bimanual manipulation. It first learns point-level action on the supporting set for different bimanual tasks, then it predicts actions on novel categories based on the foundation-model-guided affordance, enabling cross-category generalization after few-shot adaptation.
}
    \label{fig:teaser}
\end{center}
}]

\makeatletter
\renewcommand{\thefootnote}{}
\footnotetext{%
\parbox{\linewidth}{%
\raggedright
\textsuperscript{*}Equal contribution\\[-0.3em]
\textsuperscript{$\dagger$}Corresponding author\\[-0.3em]
Contact: isabella4444x@sjtu.edu.cn, linshao@nus.edu.sg.
}%
}
\renewcommand{\thefootnote}{\arabic{footnote}}
\makeatother
\begin{abstract}

Bimanual manipulation is imperative yet challenging for robots to execute complex tasks, requiring coordinated collaboration between two arms. However, existing methods for bimanual manipulation often rely on costly data collection and training, struggling to generalize to unseen objects in novel categories efficiently.
In this paper, we present Bi-Adapt, a novel framework designed for efficient generalization for bimanual manipulation via semantic correspondence. Bi-Adapt achieves cross-category affordance mapping by leveraging the strong capability of vision foundation models. Fine-tuning with restricted data on novel categories, Bi-Adapt exhibits notable generalization to out-of-category objects in a zero-shot manner.
Extensive experiments conducted in both simulation and real-world environments validate the effectiveness of our approach and demonstrate its high efficiency, achieving a high success rate on different benchmark tasks across novel categories with limited data.
Project website: \url{https://biadapt-project.github.io/}
\end{abstract}

\section{INTRODUCTION}
Bimanual manipulation is common and crucial in human-centered environments, as some tasks inherently require the collaboration of both arms. However, obtaining high-quality and efficient manipulation strategies for complex bimanual manipulation tasks remains challenging for robots due to their high dependence on coordinated arm movements and the significant variation in 3D objects across different categories. Researchers have made great advances in bimanual manipulation\cite{zhao2023dualaffordlearningcollaborativevisual, bahety2024screwmimic, mu2024robotwin}. However, some of these approaches focused only on simple tasks (e.g., `push’, `pull') for a limited number of in-category objects. They performed poorly on unseen objects outside the training set. As a result, some methods aim to solve this generalization problem through training on large-scale datasets. These data-driven methods, such as ScrewMimic~\cite{bahety2024screwmimic}, based on imitation learning (IL), require the manual collection of extensive expert demonstrations. However, collecting real-world interaction data is labor-intensive, and training on a huge dataset is time-consuming. While these works have been extensively explored in the literature, efficiently generalizing manipulation strategies for complicated bimanual tasks, particularly leveraging prior knowledge on novel categories, remains a long-standing challenge.
 
Foundation models have demonstrated exceptional performance across various fields, particularly in computer vision and natural language processing~\cite{caron2021emerging, kirillov2023segment, devlin2019bertpretrainingdeepbidirectional}.
For robotic manipulation, some works leverage the common-sense knowledge and generalization capabilities of current language and vision foundation models to improve robotic manipulation. For example, Shen et al. \cite{shen2023distilled} combine accurate 3D geometry with rich semantics from 2D foundation models by leveraging distilled feature fields, enabling few-shot language-guided manipulation that generalizes across object poses, shapes, appearances, and categories. Recent studies also suggest that the zero-shot generalization of distilled features from pre-trained foundation models holds strong power
for cross-category affordance generalization via semantic correspondence, and thus guides the manipulation on novel object categories and unseen scenarios \cite{wang2023sparsedff, ju2024robo, wang2024gendp}.
However, entirely relying on the off-the-shelf foundation models to extract affordance via semantic correspondence is not enough. The capability of obtaining positive affordance on novel categories is strictly upper-bounded by the capability of its upstream vision foundation model, the quality of images, and the accuracy of points selected.
Besides affordance, the coordinated manipulation orientation is also imperative. The bimanual manipulation strategy without adaptation for affordance and manipulation orientation could easily lead to failure when directly evaluated on unseen categories.

Most prior work focuses on learning dual-arm stabilization policies through reinforcement learning (RL) \cite{Chen2023BiDexHandsTH}. 
But these methods are often costly and not efficient, as they typically require rolling out policies many times and struggle to achieve large-scale generalization. This process is time-consuming due to the complexity of bimanual manipulation.

To overcome the limitations above, in this work, we introduce \textbf{Bi-Adapt}, a foundation-model-based framework for efficient generalization for bimanual manipulation. Bi-Adapt accumulates prior knowledge about bimanual manipulation on a constrained supporting set, and then leverages semantic correspondence from the foundation model to transfer affordance to new categories, followed by few-shot learning on a small number of instances. 
Notably, with limited robot demonstrations on novel categories, our method can demonstrate a superior performance of the finetuned model evaluated zero-shot on more instances in novel categories.

To demonstrate the performance of our framework, we conduct experiments on diverse object categories over 5 challenging tasks. Extensive experiments, including both qualitative and quantitative results, validate the effectiveness of our framework and its individual components.
In summary, our contributions are as follows:
\begin{itemize}
    \item We propose a foundation-model-based framework for generalizable bimanual manipulation across object categories for different complex tasks.
    \item We introduce a few-shot adaptation strategy following contact point selection to enhance bi-manual collaboration efficiently.
    \item Our experiments provide strong evidence supporting the effectiveness of the proposed method.
\end{itemize}
\section{Related Work}
\subsection{Bimanual Manipulation with Affordance Learning}
Bimanual manipulation offers substantial benefits over single-arm systems\cite{mu2021maniskill}. 
 Initial efforts using traditional control methods \cite{smith2012dual, mirrazavi2018unified, suarez2018design} struggled with complex models and long computation time. As a result, learning-based approaches
 have been applied to dual-arm systems \cite{chen2022towards, lin2023bi, kim2024goal},
 though challenges like the sim-to-real gap and low sample efficiency persist. The sim-to-real gap also complicates policy transfer and limits real-world deployment. 
 To address these challenges, some approaches pay attention to learning object-centric visual actionable affordance that use dense affordance maps to suggest action possibilities at every point on a 3D scan, to accelerate policy generation and benefit downstream robotic manipulation tasks \cite{wu2021vat}.
Despite these advances, affordance-learning-based methods struggle to predict precise affordance maps on unseen objects.
 In our work, we reserve the point-level action direction learning, but utilize the pre-trained vision foundation model to obtain mapped affordances on novel categories to avoid time-consuming affordance learning for bimanual manipulation.
\subsection{Foundation Models for Robotic Manipulation}
Numerous works have sought to leverage the common-sense knowledge and generalization capabilities of foundation models for robotic manipulation, mainly focusing on open-world reasoning and goal specification \cite{gu2023rt, team2024octo, kim2024openvla}.
In this work, we are interested in obtaining general affordance knowledge from existing foundation models to achieve affordance transfer. Recent advances in vision foundation models, such as DINOv2~\cite{oquab2023dinov2} and DiFT~\cite{tang2023emergent}, have demonstrated remarkable capabilities in finding semantic correspondences across objects.
Particularly, the extracted features are versatile in mapping similar points across categories.
Recent works have utilized features distilled from pre-trained foundation models for zero-shot or few-shot robotic manipulation on novel objects \cite{wang2024d3fieldsdynamic3ddescriptor}. 
However, these works mainly focused on simple grasping-based tasks with a single arm, without considering adaptation for two-arm collaboration. As the affordance predictions are strictly subjected to the foundation model's capability, the viewpoint and image quality, leading to a certain proportion of mapped points may be negative for manipulation.
In our work, we leverage the zero-shot features distilled from off-the-shelf foundation models to achieve affordance generalization via semantic correspondence across different categories. Compared to previous methods, we introduce a few-shot adaptation strategy after affordance mapping, to filter negative contact points and adjust action directions for better bimanual collaboration.
\subsection{Efficient Adaptation for Robotic Manipulation} 
Various approaches have been explored for fast adaptation~\cite{farid2021few,rakelly2019efficient,finn2017model}. Never Stop Learning~\cite{Julian2020EfficientAF} was a pioneer in combining off-policy reinforcement learning with a simple fine-tuning procedure to adapt vision-based policies to changes in background, object shape, lighting, and robot morphology. Several studies have leveraged foundation models for policy transfer~\cite{kagaya2024envbridge}, though they primarily focus on high-level adaptation rather than fine-grained policy adaptation.
AdaAfford~\cite{wang2022adaafford} provides a way for affordance adaptation requiring only test-time interactions. Where2Explore \cite{ning2024where2explore} leverages geometric similarities across categories for few-shot learning.
In our work, we introduce an efficient adaptation procedure following the contact-point selection. After the few-shot interactions on limited instances, the fine-tuned model demonstrates promising generalization to unseen instances from novel categories.
\setlength{\textfloatsep}{2.5pt}
\begin{figure}[t!]
    \centering
    \begin{subfigure}[b]{0.18\textwidth}
        \centering
        \includegraphics[width=0.95\textwidth]{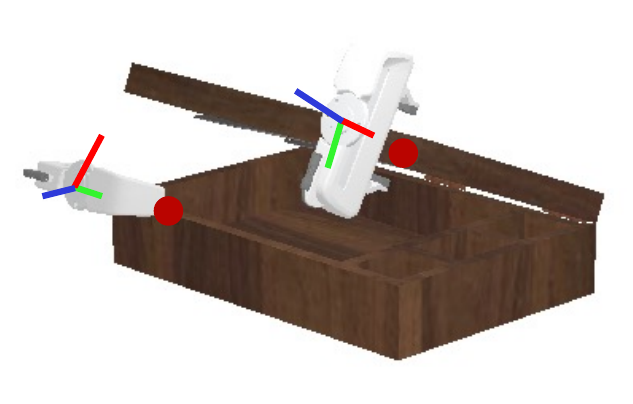}
        \caption{}
        \label{fig:a}
    \end{subfigure}
    \hspace{0.02\textwidth}
    \begin{subfigure}[b]{0.18\textwidth}
        \centering
        \includegraphics[width=0.95\textwidth]{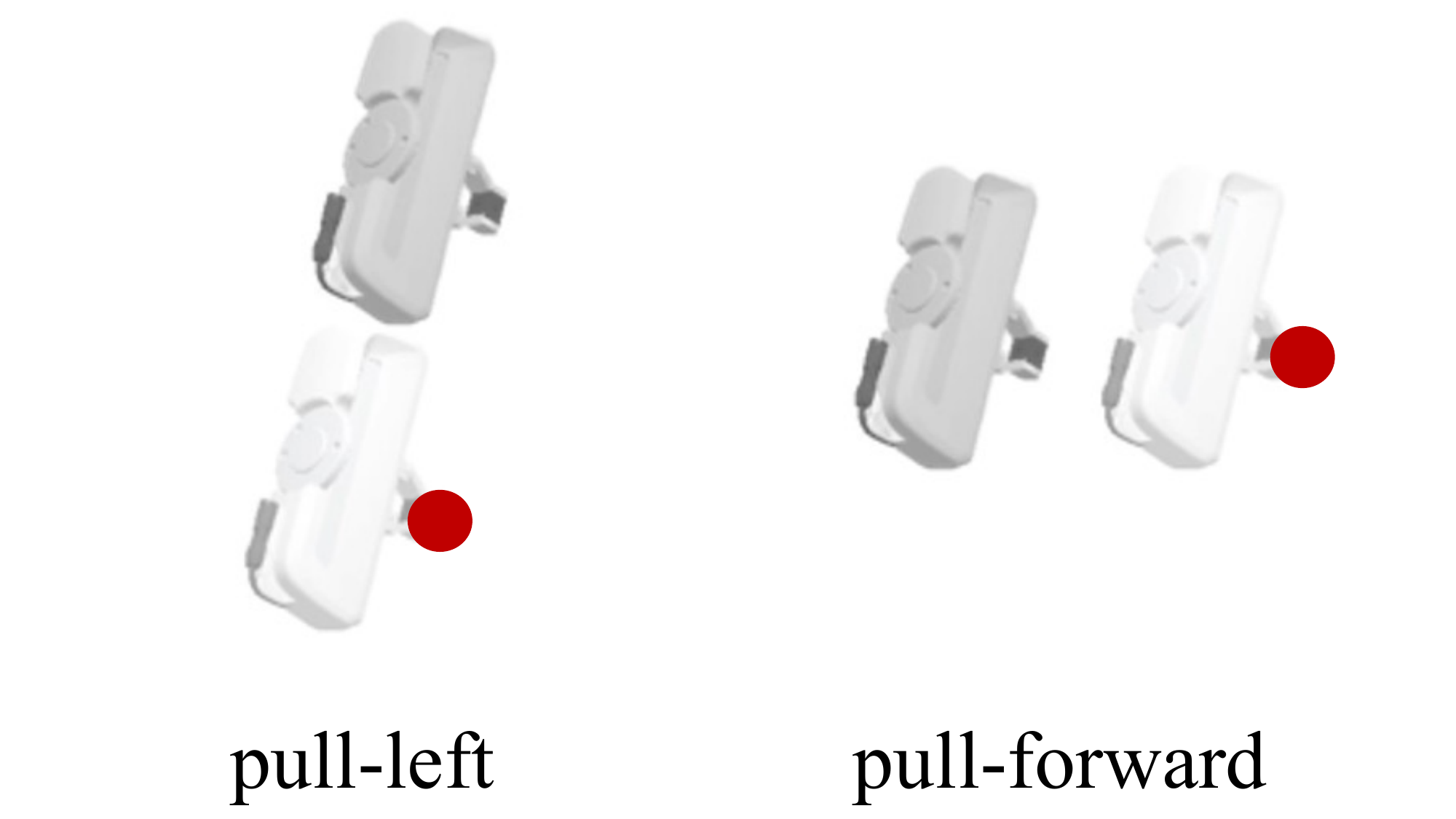}
        \caption{}
        \label{fig:b}
    \end{subfigure}
    \vskip\baselineskip
    \begin{subfigure}[b]{0.5\textwidth}
        \centering
        \includegraphics[width=0.95\linewidth]{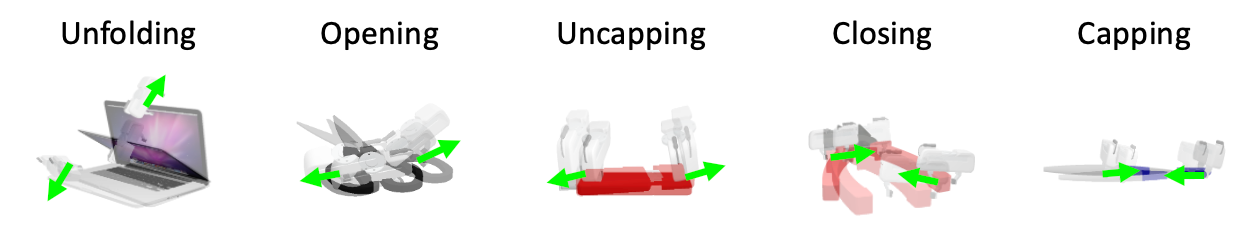}
        \label{fig:c}
        \vspace{-3mm}
        \caption{}
    \end{subfigure}
    \caption{(a) Simulation environment with gripper frames (red, green, blue axes for forward, leftward, upward); (b)
Two \(SE(3)\)-parametrized action primitives, visualized by two key frames with red contact points and time steps from transparent to solid grippers; (c) Action definitions of different tasks.}
    \label{fig:sim_actiondir}
\end{figure}
\begin{figure*}[t!]
    \includegraphics[width=0.95\linewidth]{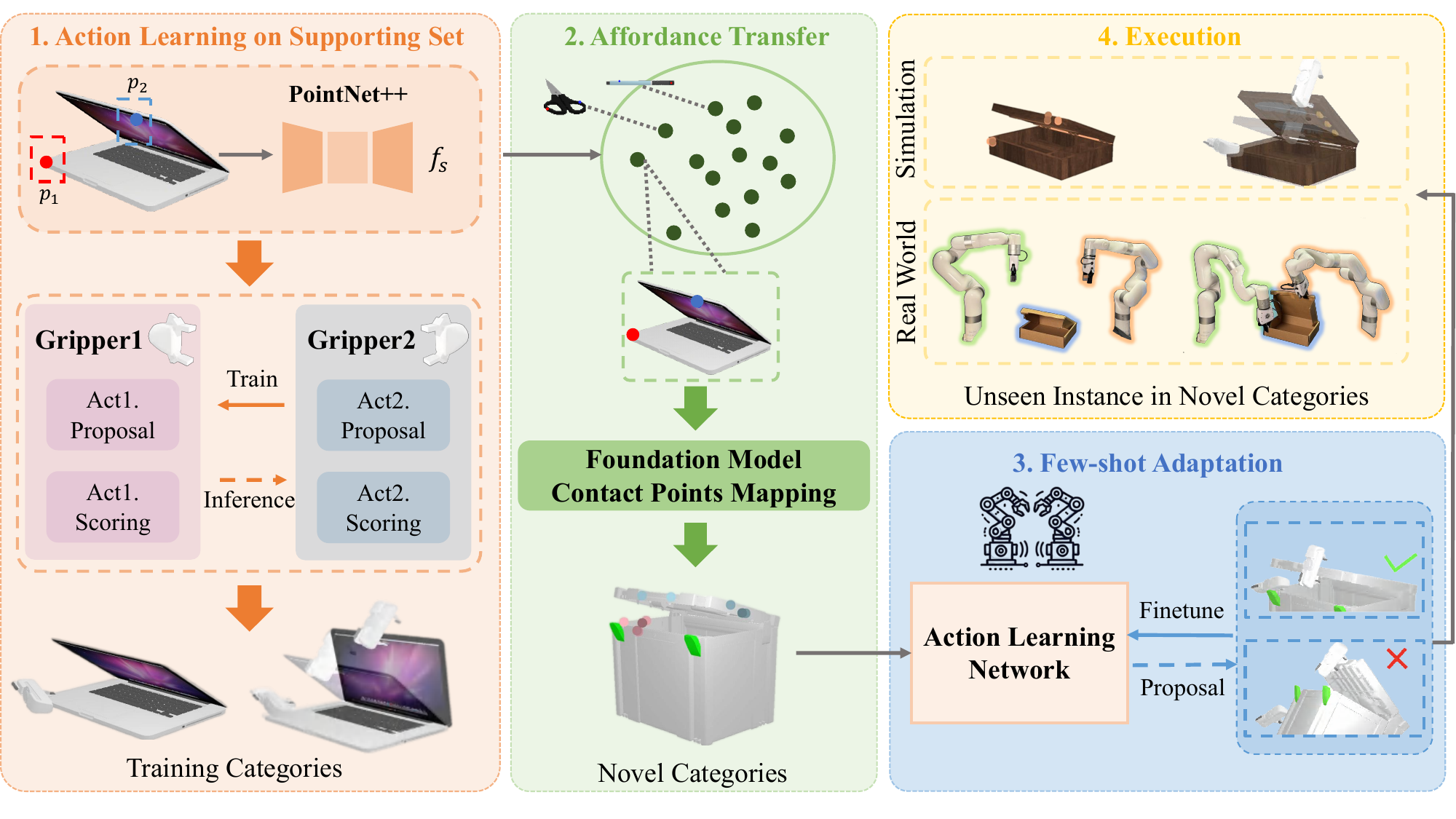}
    \centering
    \caption{\textbf{Pipeline.} \textbf{(Left)} We first train the Action Learning Network on the supporting set. The corresponding affordance and action distribution serve as our prior knowledge. \textbf{(Middle)} Then, we make contact points mapping from training categories to novel categories, leveraging the foundation model. \textbf{(Lower Right)} After that, the pre-trained network proposes actions based on mapped contact-point pairs on novel categories and fine-tuned with the interaction results. \textbf{(Upper Right)} Finally, the fine-tuned networks can facilitate manipulating unseen instances from novel categories with better performance.}
    \label{fig:pipeline}
\end{figure*}
\section{Problem Formulation}
\subsection{General Setting} We place a 3D object on the ground, given its partially scanned point cloud observation \(O \in \mathbb{R}^{ N\times 3} \) and a task \( T \), where \(N\) denotes the number of point clouds. The framework is required to propose two grippers' actions \( u_1 = (p_1, R_1)\) and \( u_2 = (p_2, R_2)\), in which \(p_i \in O\) is a contact point and \(R_i \in SO(3) \) is the gripper's orientation. Each action \(u_i\) includes contacting at point \(p_i\) and pulling to a specific direction without changing the orientation.
\Cref{fig:sim_actiondir} shows our simulation environment, two action primitives, and presents the action definitions of different tasks. 
\subsection{Task Formulation}
We formulate five benchmark tasks and set success judgments for them.
For all tasks, success requires the movement distance of the target joint to be less than a set threshold while the object does not fall or flip.
Here are the task-specific criteria:
\begin{itemize}
    \item \textbf{Unfolding, Opening, Closing}: The rotational joint angle change must exceed 0.10 times the joint's range.
    \item \textbf{Uncapping, Capping}: Two parts of the prismatic joint are separated or drawn to each other by over 0.05\,\text{m}.
\end{itemize}

\section{Method}
Our goal is to achieve bimanual manipulation tasks on novel categories that efficiently learn from prior knowledge. 
\textbf{Method Overview.} 
\label{method:overview}
\Cref{fig:pipeline} shows our pipeline. First, we begin by learning bimanual point-level manipulation orientation on training categories to construct a supporting set (\Cref{method:pretrain}). Then, we transfer affordance from the training categories in the supporting set to novel categories by using the foundation model to map contact points (\Cref{sec:affordance_transfer}). After few-shot adaptation on novel categories (\Cref{sec:fintune}),  the fine-tuned networks improve performance on unseen instances in novel categories. \Cref{sec:opt} describes network architectures and the training strategy.

\subsection{Action Learning for Building Supporting Set} 
\label{method:pretrain}
To conduct cross-category few-shot adaptation tasks, we first need to build a supporting set that captures key affordances and action patterns from known categories. This supporting set serves as a knowledge prior, enabling the transfer of learned manipulation strategies to a broader range of novel object categories.

Visual manipulation affordance has fine-grained manipulation information and is capable of generalizing to unseen objects within the same category. Building upon previous works \cite{mo2021where2act, zhao2023dualaffordlearningcollaborativevisual}, we design a Perception Module to learn collaborative visual actionable affordance and interaction policy for dual-gripper manipulation tasks over diverse objects in the supporting set. Given the vast combinatorial action space of two grippers, we mitigate the complexity by disentangling composite actions into two sequentially conditioned actions. Specifically, we design two coupled submodules in the Perception Module: the First Gripper Module  $\mathcal{M}_1$(left), and the Second Gipper Module $\mathcal{M}_2$ (right), where $\mathcal{M}_1$ proposes $u_1$ and $\mathcal{M}_2$ proposes $u_2$ based on $u_1$. 
As indicated by the arrows in \Cref{fig:pipeline}, the training and inference procedures share the same architecture but follow opposite dataflow directions. For inference, the dataflow direction is intuitive: $\mathcal{M}_1$ proposes \(u_1\), and $\mathcal{M}_2$ proposes \(u_2\) conditioned on \(u_1\). But this approach does not ensure that the \(u_1\) is optimal for collaboration. To address this, we adopt a reversed dataflow strategy during training: $\mathcal{M}_2$ is trained first, followed by $\mathcal{M}_1$, which learns based on $\mathcal{M}_2$. Specifically, $\mathcal{M}_2$ is trained to generate suitable \(u_2\) given the diverse \(u_1\) samples from the training dataset. Once $\mathcal{M}_2$ is capable of generating \(u_2\) that effectively collaborate with various \(u_1\), $\mathcal{M}_1$ is trained to propose \(u_1\) that facilitate successful collaboration with $\mathcal{M}_2$.

Each gripper module consists of two networks: the Action Proposal Network $\mathcal{A}$ and the Action Scoring Network $\mathcal{C}$.  
Given the contact point, $\mathcal{A}$ predicts the gripper orientation, while $\mathcal{C}$ evaluates whether the proposed action is suitable for manipulation. We detail the design of networks and training strategy in \Cref{sec:opt}.
However, this series of networks suffers significant drops in novel categories. To enable broader generalization, we leverage semantic correspondence to link different categories.
\subsection{Affordance Transfer}
\label{sec:affordance_transfer}
When encountering an unfamiliar object, humans often recall experiences to determine how to interact with it. Inspired by this, we propose a similar approach to robots. Objects from different categories often share geometric similarities and semantic correspondence. For example, \textit{Scissors} and \textit{Pliers} have similar shapes and can be operated using similar actions. If a robot has mastered the skill of manipulating the Scissors,  it can leverage this knowledge to handle the Pliers. The cross-category semantic correspondence of affordance can serve as a bridge to connect them. In our study, we use the foundation model to map contact points from objects in our supporting set to novel objects, guiding the manipulation.
\subsubsection{Affordance Representation and Collection}
We define the affordance as the contact points on the object at the first frame when this contact takes place. For two grippers, we select one point for each, $p_1, p_2 \in O$.

When collecting data on training categories, we capture the RGBD image before the interaction and record the contact points. We only utilize the contact points with successful outcomes for affordance transferring. 

\subsubsection{Contact Points Mapping via Semantic Correspondence}
\label{mapping}
When encountering an unseen object and a task, we first retrieve source objects within the same task in the supporting set. With the recorded contact points, we can sample multiple successful cases as our source data.
Then we transfer the \(2D\) affordance from the source objects to the target object. In our study, we utilize the emergent semantic correspondence ability from the foundation model to map the retrieved contact points to the new object.

Given a source image $I_s$ with two contact points $p_{s_1}^{2D}$, $p_{s_2}^{2D}$, a target image $I_t$, we aim to find the corresponding points $p_{t_1}^{2D}$, $p_{t_2}^{2D}$ in $I_t$. With the foundation model, we extract the diffusion features (DIFT) of $I_s$ and $I_t$ as the steps shown in \cite{tang2023emergent}. By adding noise to $I_t$ first and denoising through the diffusion process next, the diffusion features are generated.  As the diffusion features correspond to each pixel in $I_t$, with each source contact point $p_{s}^{2D}$ and its diffusion feature $f_{s}$ obtained, we can calculate the similarity in the form of:
\begin{equation}
  \mathit{similarity} = \cos{f_{p_{s}^{2D}}} \cdot \cos{f_{p_{t^j}^{2D}}}
  \label{eq:similarity}
\end{equation}
for each pixel $p_{t^j}^{2D}$ in $I_t$. 
Then we can find the pixel $p_{t_i}^{2D}$ with the highest similarity to the source point $p_{s_i}^{2D}$ in the diffusion features of the target image $I_t$. Based on the partial point cloud of the target object generated from the depth map $D$, we back-project the pixel $p_{t_i}^{2D}$ to get the 3D contact point $p_{t_i}^{3D}$. 
Since the resulting contact points may not be suitable for manipulation on a specific task, we use multiple source images and their contact point pairs to obtain multiple contact-point pairs candidates, for further selection in \Cref{sec:fintune}.
\begin{figure*}[t!]
    \centering
    \begin{subfigure}[b]{0.5\textwidth}
        \centering
        \includegraphics[width=\textwidth]{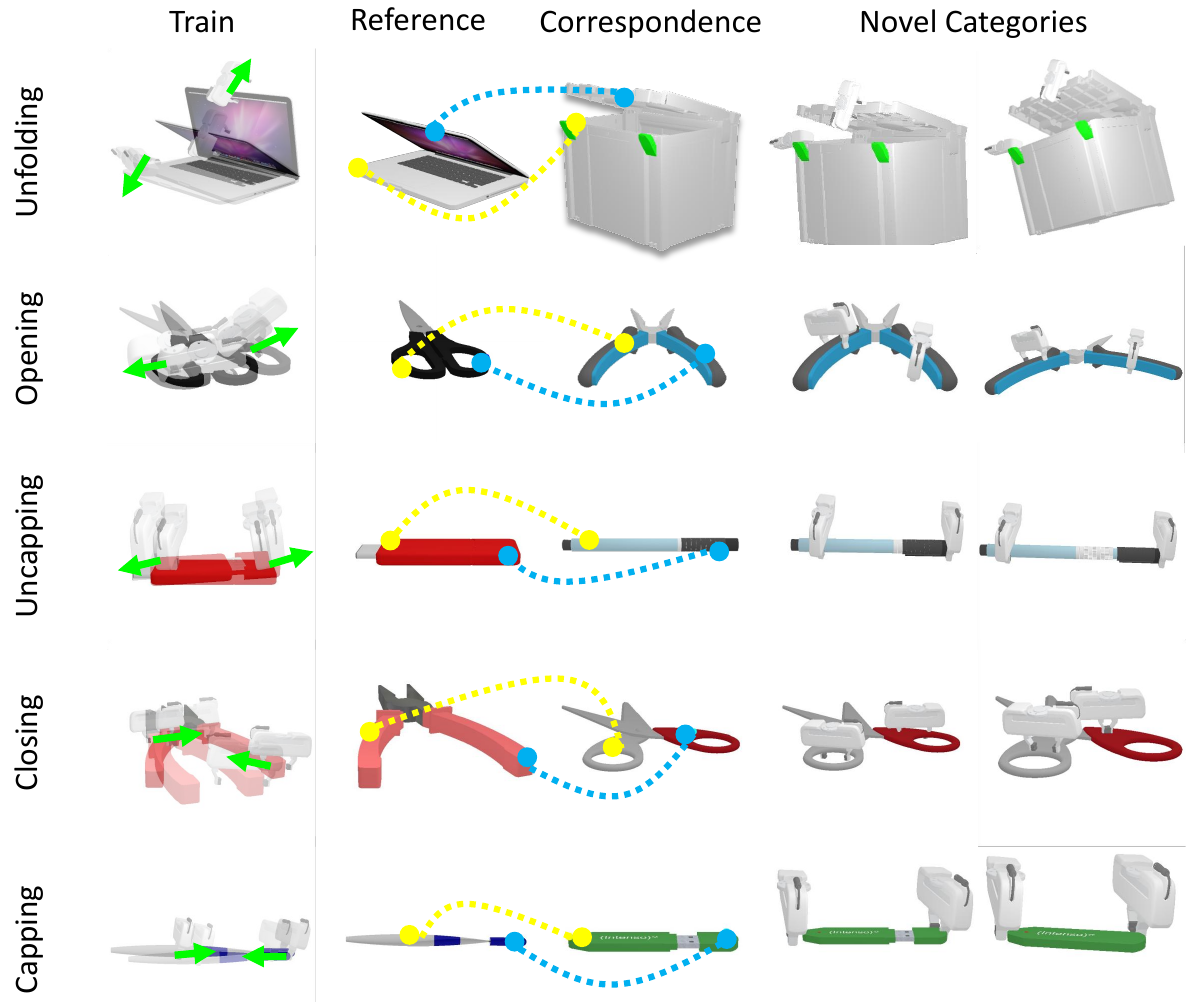}
        \label{fig:manip_results}
    \end{subfigure}
    \hspace{0.02\textwidth}
    \tikz \draw[dashed, thick] (0,-0.2) -- (0,7.5);
    \hspace{0.02\textwidth}
    \begin{subfigure}[b]{0.32\textwidth}
        \centering
        \includegraphics[width=\textwidth]{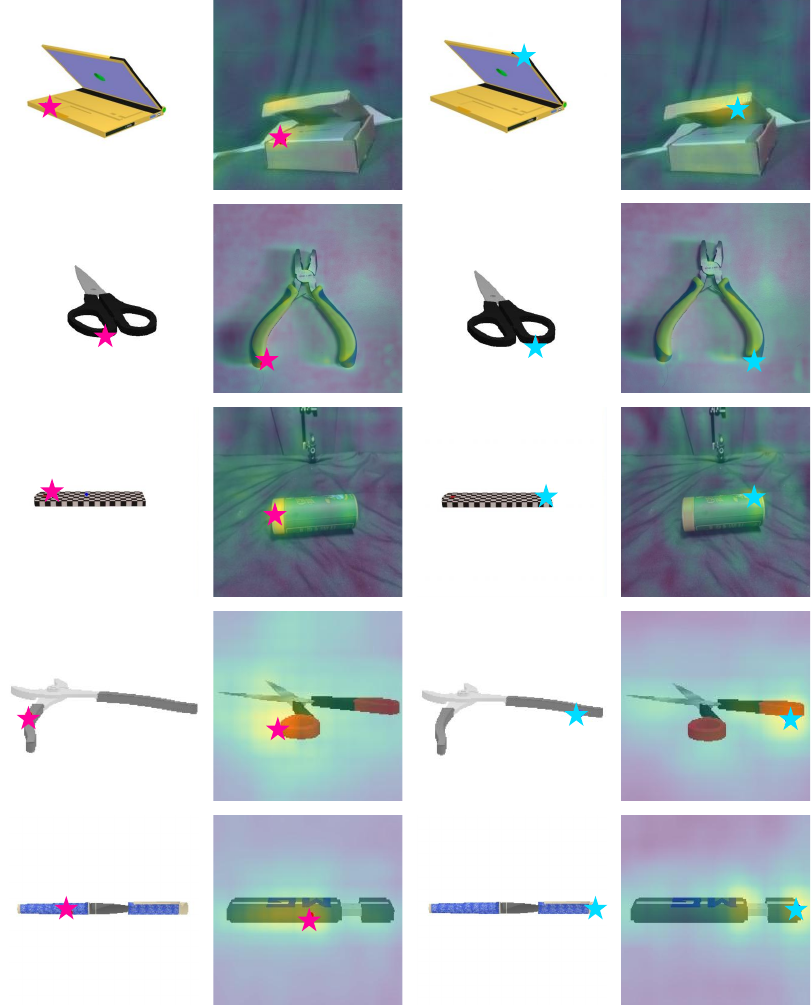}
        \label{fig:dift}
    \end{subfigure}
    
     \caption{(\textbf{Left}) Manipulation results across tasks. (\textbf{Right}) Cross-category affordance generalization. Each group shows a source image with contact points (\textcolor{magenta}{\ding{72}}: first gripper, \textcolor{cyan}{\ding{72}}: second gripper) and a target image with inferred points. Highlight intensity indicates correspondence. }
    \label{fig:manip_results_diffusion}
    \vspace{-3mm}
\end{figure*}
\subsection{Few-shot Adaptation}
\label{sec:fintune}
Not all contact point candidates are suitable for manipulation, as mentioned in \Cref{mapping}. And the actions proposed by the previous Perception Module can easily lead to failure due to the significant variance in physical properties and geometries among different categories. To improve the performance on novel categories, we introduce the few-shot adaptation procedure, in which the networks are finetuned with the results obtained by executing proposed actions.

As shown in \Cref{fig:pipeline}, the pre-trained Perception Module predicts two grippers' orientations based on sampled contact-point pair candidates. The actions with the greatest likelihood of success are executed simultaneously, with the outcomes for updating the networks. 
Therefore, the Perception Module can better determine whether the proposed actions succeed and adjust them to fit the novel categories.

Through just a few interactions on limited instances in the novel category, the Perception Module explores a novel category's semantic significant areas selected by the foundation model, and better understands the geometrical and physical difference compared to learned categories. After the few-shot adaptation, the knowledge about the manipulation is transferred to novel categories. Thanks to the generalization of geometrical characteristics and physical properties within one category, our adapted module could manipulate unseen objects from this category without additional interactions.
\subsection{Network Architecture and Training Strategy}
\label{sec:opt}
\qheading{Backbone Feature Extractors.} The networks in the Perception Module receive three kinds of input entities or intermediate results: point cloud \(O\),  contact point \(p\), and gripper orientation \(R\). In different submodules, the backbone feature extractors share the same architectures. We use a segmentation-version PointNet++ (Qi et al., 2017 \cite{qi2017pointnet++}) to extract per-point feature \(f_s\in \mathbb{R}^{128} \) from \(O\), and employ three MLP networks to respectively encode  \(p\) and \(R\) into  \(f_p\in \mathbb{R}^{32} \) and \(f_R\in \mathbb{R}^{32} \).

\qheading{Action Scoring Loss.} For $\mathcal{C}_2$, we use ground-truth results to supervise it. Given the interaction data with the ground-truth result \(r\), where \(r = 1\) means positive and \(r = 0\) means negative, we can train $\mathcal{C}_2$ using the standard binary cross-entropy loss, in the form of:
\begin{equation}
    \mathcal{L}_{\mathcal{C}_2} = r_i \log(\mathcal{C}_2(f^{in}_{\mathcal{C}_2}) + (1-r_i)\log(1-\mathcal{C}_2(f^{in}_{\mathcal{C}_2})).
  \label{eq:loss_c2}
\end{equation}
We evaluate the first gripper's action by evaluating its potential for the second gripper to collaborate with.
\qheading{Action Proposal Loss.} $\mathcal{A}_1$ and $\mathcal{A}_2$ are implemented as cVAE \cite{sohn2015learning}. For the \(j_{th}\) gripper, we adopt geodesic distance loss to measure the error between the reconstructed orientation $\hat{R}_j$ and ground-truth $R_j$, and KL Divergence to measure the difference between two distributions:
\begin{equation}
   \mathcal{L}_{A_j} = \mathcal{L}_{geo}(\hat{R}_j,R_j) + D_{KL}(q(z\mid \hat{R}_j,f^{in}_{\mathcal{A}_j})\parallel \mathcal{N}(0,1)).
  \label{eq: loss a}
\end{equation}
For training Action Scoring Network $\mathcal{C}_1$ and $\mathcal{C}_2$, we balance the portion of successful cases and failed cases equally to ensure it can critically score. For training Action Proposal Network $\mathcal{A}_1$ and $\mathcal{A}_2$, we only use successful data for training.
For training categories, we generate offline interaction data with multiple processes for training. For few-shot learning on novel categories, we sample affordances and propose actions to execute, collecting the online results to fine-tune networks.
\begin{table}[t!]
  \centering
  \caption{Baseline comparisons of sample success rate (\%).}
  \scriptsize
  \setlength{\tabcolsep}{3.7pt}  
  \begin{tabular}{lccccc}
    \toprule
    \multicolumn{6}{c}{\textbf{Unseen instances in novel categories}} \\
    \midrule
    Method & Unfolding & Opening & Uncapping  & Closing & Capping \\
    \midrule
    Heuristic     & 24.70  & 19.82  & 31.33   & 33.74   & 43.20 \\
    M-Where2Act~\cite{mo2021where2act} & 32.40 & 20.66& 29.10& 18.45 & 17.30 \\
    DualAfford~\cite{zhao2023dualaffordlearningcollaborativevisual}    & 33.50  & 21.90  & 19.60   & 25.50    & 35.00 \\
    \cellcolor{lightblue}\textbf{Ours} & \cellcolor{lightblue}\textbf{70.00} & \cellcolor{lightblue}\textbf{67.00} & \cellcolor{lightblue}\textbf{61.62}  &  \cellcolor{lightblue}\textbf{61.12} & \cellcolor{lightblue}\textbf{59.00} \\
    \bottomrule
  \end{tabular}
  \label{tab:baseline}
\end{table}
\vspace*{-3mm}
\section{Experiments}
\label{sec:experiments}
To evaluate our framework’s ability, we seek to answer the following research questions:

Q1: How successful are our proposed actions on unseen objects in novel categories, and how does it compare to other baselines for bimanual manipulation tasks? (\Cref{overall_performance})

Q2: How's the cross-category affordance generalization ability via semantic correspondence, and how does it compare to the prior method without using the foundation model? (\Cref{ablation_AT})

Q3: How's the effectiveness of the few-shot adaptation strategy, and does it improve the performance compared with no adaptation? (\Cref{ablation_FA})

Q4: How efficient is few-shot adaptation, and how much data is required for fine-tuning? (\Cref{sec:efficiency})
\begin{table*}[t!]
  \centering
  \caption{Ablation studies on sample success rate (\%). Training categories and novel categories (seen/unseen instances).}
  \scriptsize
  \begin{tabular}{l|c|cc|c|cc|c|cc|c|cc|c|cc}
    \toprule
    \multirow{2}{*}{Method} 
      & \multicolumn{3}{c|}{Unfolding} 
      & \multicolumn{3}{c|}{Opening} 
      & \multicolumn{3}{c|}{Uncapping} 
      & \multicolumn{3}{c|}{Closing} 
      & \multicolumn{3}{c}{Capping} \\
    \cmidrule(lr){2-4}\cmidrule(lr){5-7}\cmidrule(lr){8-10}\cmidrule(lr){11-13}\cmidrule(lr){14-16}
      \cmidrule(lr){2-4}\cmidrule(lr){5-7}\cmidrule(lr){8-10}\cmidrule(lr){11-13}\cmidrule(lr){14-16}
      & \multirow{2}{*}{Train} & \multicolumn{2}{c|}{Novel} 
      & \multirow{2}{*}{Train} & \multicolumn{2}{c|}{Novel} 
      & \multirow{2}{*}{Train} & \multicolumn{2}{c|}{Novel} 
      & \multirow{2}{*}{Train} & \multicolumn{2}{c|}{Novel} 
      & \multirow{2}{*}{Train} & \multicolumn{2}{c}{Novel} \\
    \cmidrule(lr){3-4}\cmidrule(lr){6-7}\cmidrule(lr){9-10}\cmidrule(lr){12-13}\cmidrule(lr){15-16}
      &  & Seen & Unseen 
      &  & Seen & Unseen 
      &  & Seen & Unseen 
      &  & Seen & Unseen 
      &  & Seen & Unseen \\
    \midrule
    Ours w/o AT 
      & \multirow{3}{*}{75.50} & 47.95 & 37.76
      & \multirow{3}{*}{63.50} & 32.61 & 31.23
      & \multirow{3}{*}{43.22} & 41.30 & 21.71
      & \multirow{3}{*}{76.00} &  39.29 & 38.70
      & \multirow{3}{*}{75.00} & 39.13 & 36.00 \\
    Ours w/o FA 
      &   & 62.50 & 62.96
      &   & 61.90 & 47.83
      &   & 53.19 & 52.94
      &   & 45.65 & 48.72
      &   & 59.52 & 57.58 \\
    \cellcolor{lightblue}\textbf{Ours} 
      &   & \cellcolor{lightblue}\textbf{68.00} & \cellcolor{lightblue}\textbf{70.00}
      &   & \cellcolor{lightblue}\textbf{72.00} & \cellcolor{lightblue}\textbf{67.00}
      &   & \cellcolor{lightblue}\textbf{63.00} & \cellcolor{lightblue}\textbf{61.62}
      &   & \cellcolor{lightblue}\textbf{56.00} & \cellcolor{lightblue}\textbf{61.12}
      &   & \cellcolor{lightblue}\textbf{64.00} & \cellcolor{lightblue}\textbf{59.00} \\
    \bottomrule
  \end{tabular}
  \label{tab:ablations}
\end{table*}
\begin{figure}[t!]
    \centering
    \includegraphics[width=0.95\linewidth]{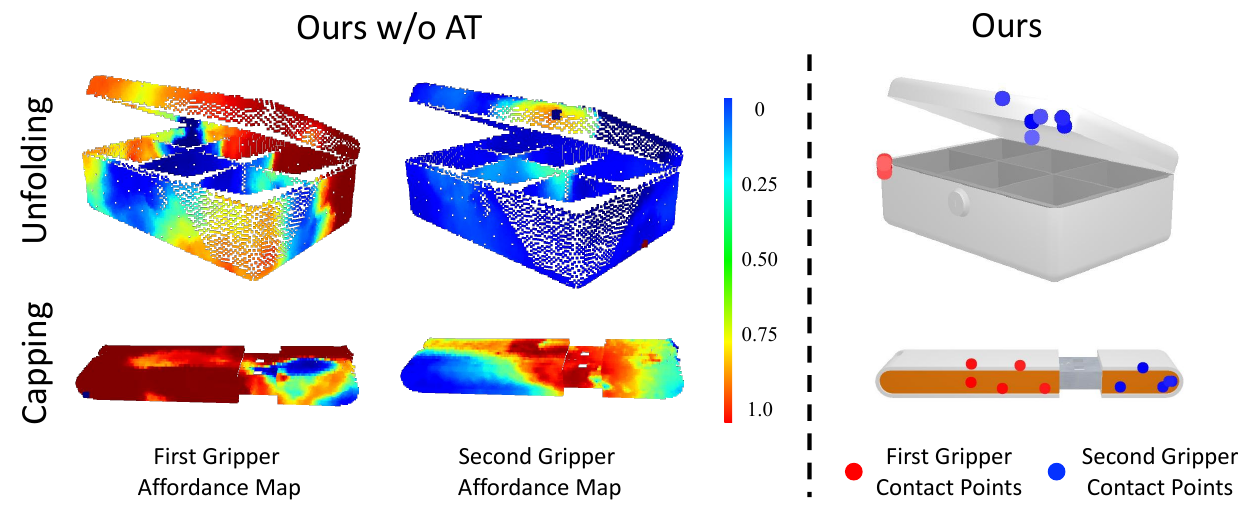}
    \caption{Visualization of affordance comparison. }
    \label{fig:affordance}
    \vspace{-2mm}
\end{figure}
\subsection{Environment Settings and Data}
\label{Environment Settings and Data}
 \qheading{Environment Settings.}Following Dualafford \cite{ zhao2023dualaffordlearningcollaborativevisual}, we use SAPIEN \cite{xiang2020sapien} with NVIDIA PhysX for simulation, and we use two Franka Panda Flying grippers as robot actuators. 
 
  \qheading{Data.}We conduct experiments with the large-scale PartNet-Mobility \cite{Mo_2019_CVPR} and ShapeNet \cite{chang2015shapenet} dataset. 
 We use the terms \textbf{training categories} and \textbf{novel categories} to denote whether objects of the same category are presented in the supporting set. In the \textbf{novel categories},
 a smaller proportion (less than 30\%) of instances that are exclusively used for few-shot learning are called \textbf{seen instances}, while many more unseen instances are only used for evaluation and are called \textbf{unseen instances}.
 In summary, we use 61 articulated objects spanning 6 categories. 
\subsection{Evaluation Metrics}
\qheading{Sample success rate.}
We use the sample success rate to assess the method's ability to propose successful actions. 
Notably, we compare our method using one budget of interaction data (50 demonstrations on novel categories) to compare with other baselines and ablation versions, shown in \Cref{tab:baseline} and \Cref{tab:ablations}.
\subsection{Overall Performance}
\label{overall_performance}
\textbf{Baselines.} To answer Q1, we compare our framework with several baselines, as shown in \Cref{tab:baseline}. This comparison includes diverse methods to address the challenge of cross-category generalization for bimanual manipulation. They were evaluated on unseen instances in novel categories over 5 complex bimanual manipulation tasks. 

\textbf{Heuristic} is a manually-scripted approach, where we acquire the ground-truth object poses and hand-engineer a set of rules for different tasks. For example, in the uncapping task, two contact points are sampled on each part of the prismatic joint, and the grippers are oriented to pull in opposite directions.
\textbf{M-Where2Act} is a dual-gripper version of Where2Act \cite{mo2021where2act} approach. While Where2Act originally considers interactions for a single gripper, we first train a separate model for each gripper and then combine them without considering collaboration. \textbf{DualAfford \cite{zhao2023dualaffordlearningcollaborativevisual}} is a dual-gripper affordance learning method considering two arms' actions as a combination. Unlike \textbf{M-Where2Act}, it models the action of one gripper conditioned on the action of the other gripper, enabling coordinated dual-arm manipulation. This approach serves as our baseline for affordance transferring and few-shot adaptation. 
Experiments demonstrate that our method outperforms all baselines regarding success rate.
\Cref{fig:manip_results} illustrates the manipulation results generated from our method.
The \textbf{Heuristic} baseline shows a low success rate since the manually engineered rules are difficult to fit all objects with highly varied geometries of shapes.
 Moreover, this approach is time-consuming and labor-intensive, and impractical to generalize. 
The \textbf{M-Where2Act} approach struggles with tasks strongly demanding two arms' coordination.
Although \textbf{DualAfford} achieves relatively high performance on training categories, its performance drops dramatically when encountering unseen objects from novel categories. The long training procedure with a large amount of data is also costly. 
Our work employs a similar bimanual-conditioned action learning method as prior knowledge.

To answer Q2 and Q3, we compare to ablated versions of our framework to verify each component's effectiveness:
\textbf{Ours w/o AT} is our method without transferring affordance based on semantic correspondence, where the Perception Module, after finetuning with 50 demonstrations of novel categories(the same as our full version's few-shot adaptation strategy), and then evaluated on unseen instances.
\textbf{Ours w/o FA} is our method without a few-shot adaptation strategy, where the pre-trained Perception Module is directly evaluated on unseen instances with all mapped contact points via semantic correspondence.
\Cref{tab:ablations} clearly shows that each component improves our framework’s performance. We detail the analysis in \Cref{ablation_AT} and \Cref{ablation_FA}.

\begin{figure}[t!]
    \centering
    \includegraphics[width=0.95\linewidth]{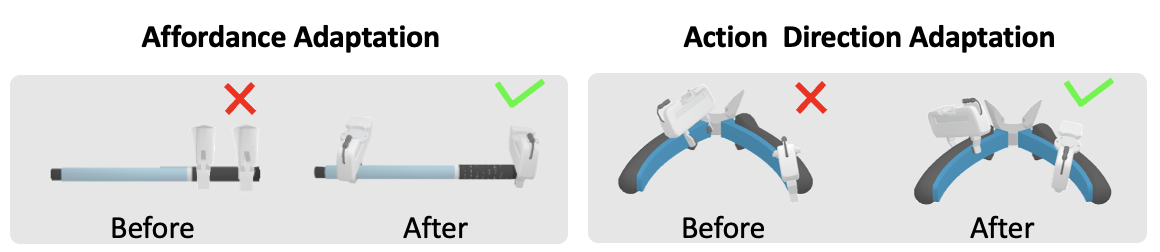}
    \caption{Visualization of the adaptation results.}
    \label{fig:adaptation}
\end{figure}
\begin{figure*}[t!]
    \centering
    \includegraphics[width=0.95\textwidth]{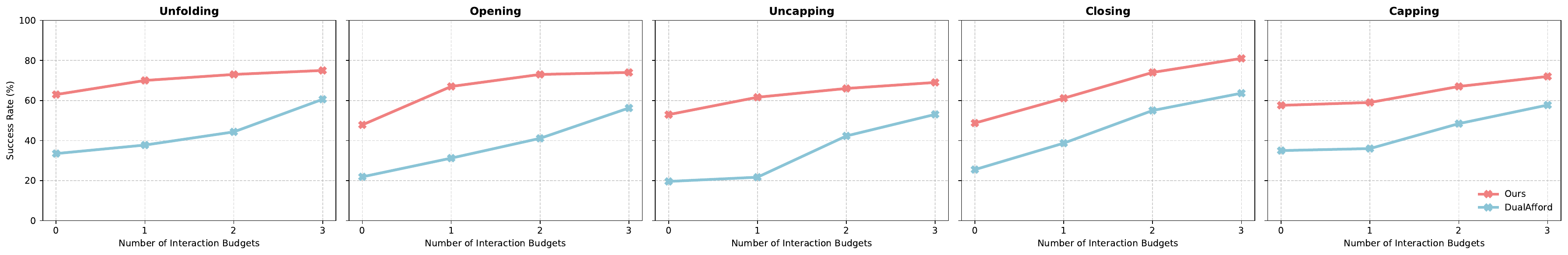}
    \caption{Efficiency comparison of Ours and DualAfford with varying novel-category interaction data on unseen instances.}  
    \label{fig:adapt_change}
    \vspace{-3mm}
\end{figure*}
\begin{figure}[t!]
    \centering
    \begin{subfigure}[b]{0.22\textwidth}
        \centering
        \includegraphics[width=\textwidth]{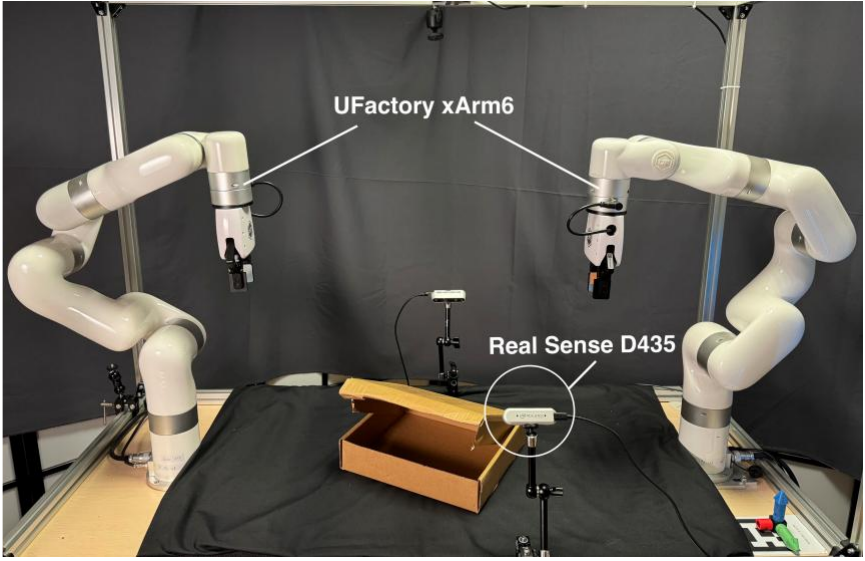}
        \caption{Robot Setup}
        \label{fig:real_robot}
    \end{subfigure}
    \hspace{0.01\textwidth}
    \begin{subfigure}[b]{0.2\textwidth}
        \centering
        \includegraphics[width=\textwidth]{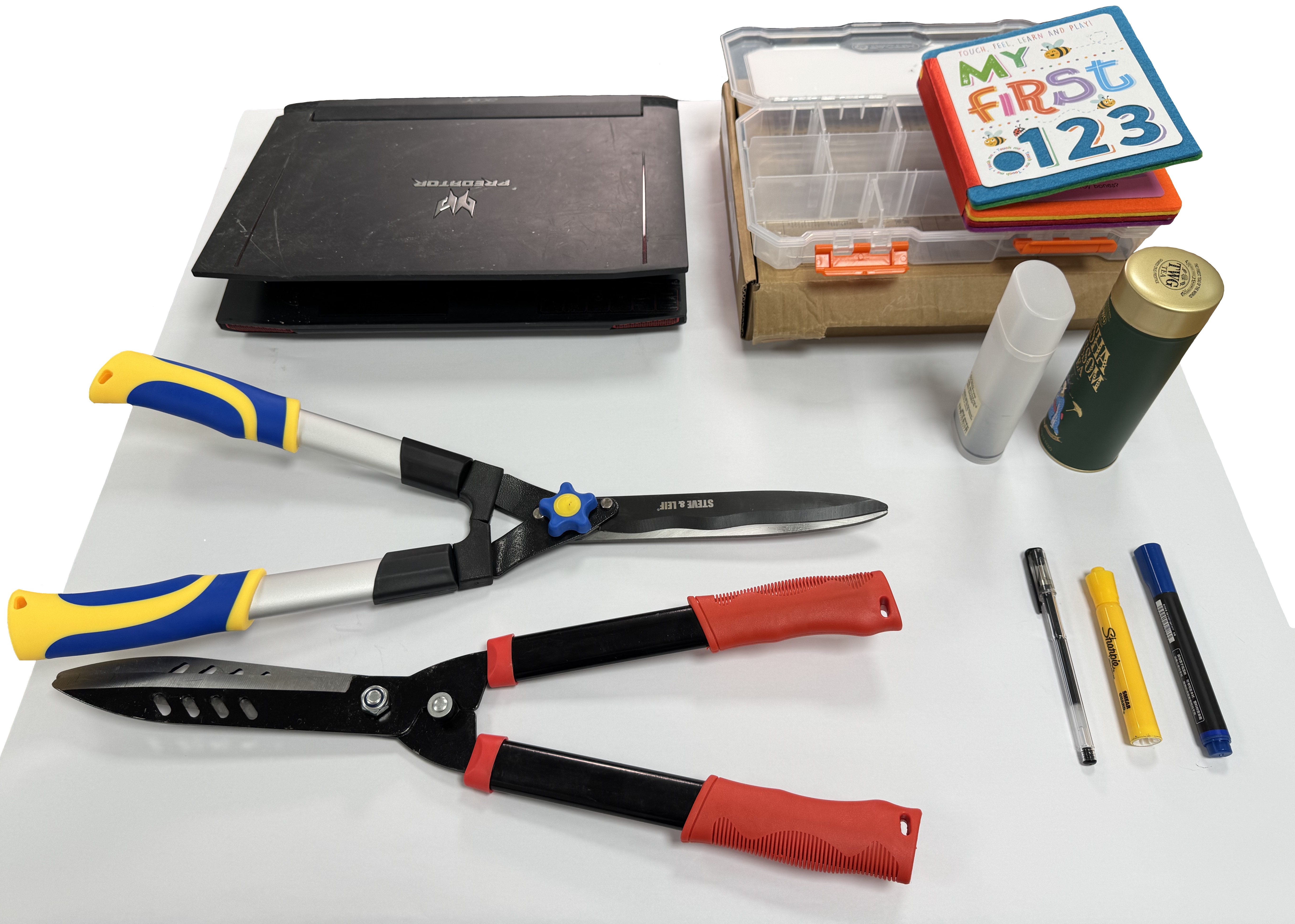}
        \caption{Objects}
        \label{fig:real_obj}
    \end{subfigure}
    \vspace{-2mm}
    \vskip\baselineskip
    \begin{subfigure}[b]{0.5\textwidth}
        \centering
        \includegraphics[width=0.95\linewidth]{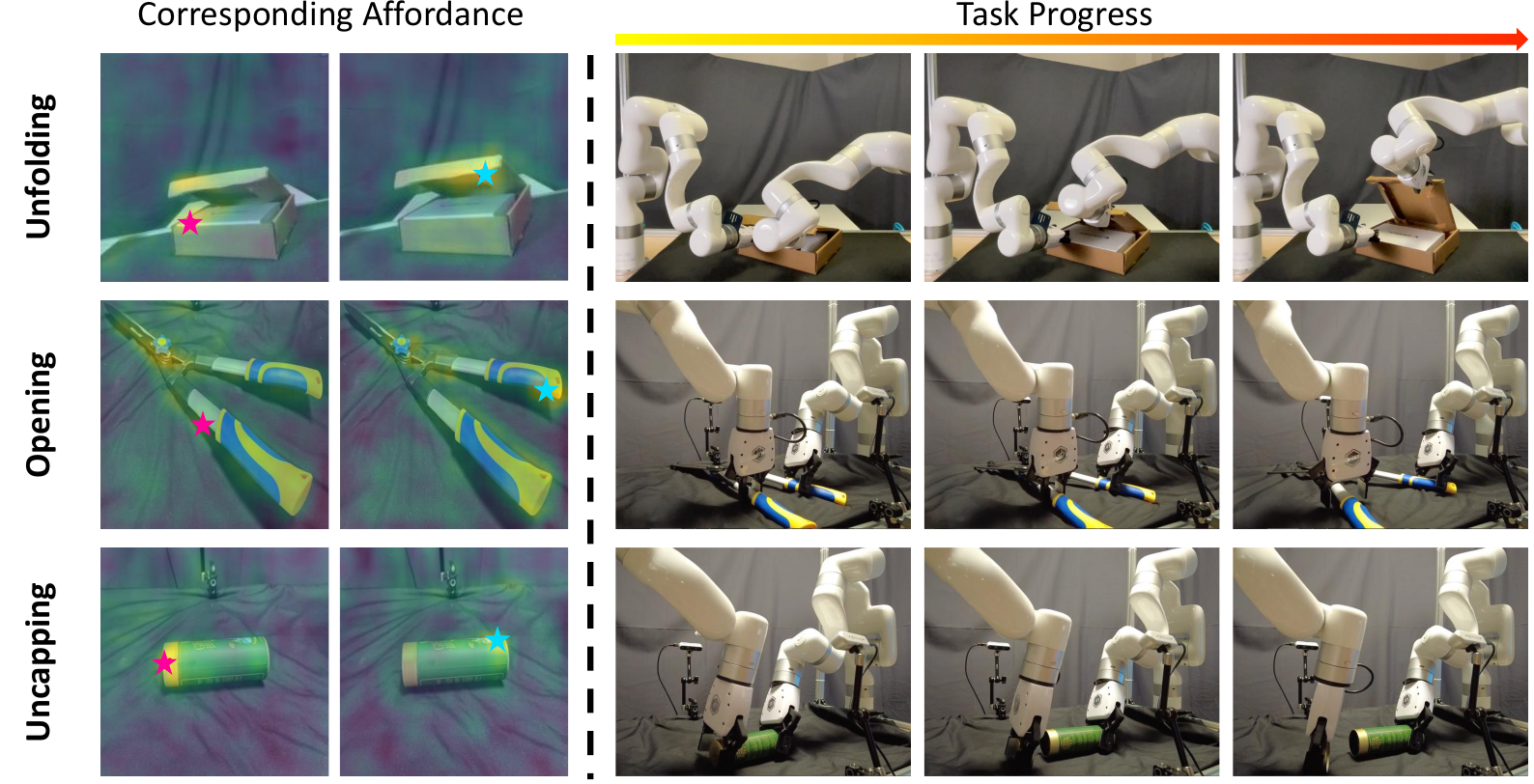}
        \caption{Real-world tasks. Each row represents a task. From left to right, we respectively show the corresponding affordance on novel categories (\textcolor{magenta}{\ding{72}}: first gripper, \textcolor{cyan}{\ding{72}}: second gripper) and the manipulation progress.}
        \label{fig:real_manip}
    \end{subfigure}

        \caption{Real-world Experiments.} 
    \label{fig:real_setting}
\end{figure}

\subsection{Cross-category Affordance Generalization}
\label{ablation_AT}
\Cref{fig:dift} visualizes the affordance mapping results via semantic correspondence, illustrating the foundation model's strong power in cross-category affordance generalization.
As shown in 
\Cref{fig:affordance}, the affordance map generated from \textbf{Ours w/o AT} for novel categories is much less accurate than our method's contact-point-pair candidates. 

Compared with \textbf{Ours w/o AT}, we observe that the semantic correspondence could act as an essential guidance for the cross-category affordance generalization, and thus significantly improves the overall performance (Q2).
\subsection{Effectiveness of Adaptation}
\label{ablation_FA}
Compared with \textbf{Ours w/o FA}, which directly uses all contact points mapped by the foundation model and directions predicted by pre-trained networks, we find that this strategy can help filter negative contact points and adjust action directions to suit new categories.

As shown in \Cref{fig:adaptation}, after adaptation, the contact points selected and each gripper's orientation both tend to be more suitable for manipulation on novel categories, addressing Q3.

As discussed in \Cref{sec:affordance_transfer}, we try to find multiple contact point pairs through semantic correspondence approaches when encountering an unseen object, and these multiple pairs serve as candidates for sampling and selection. Since actions proposed by the pre-trained networks can lead to failure on unseen instances, the adaptation procedure adaptation eliminates negative candidates and inaccurate action directions.
\subsection{Effeciency}
\label{sec:efficiency}
To answer Q4, we evaluate the efficiency of our method on few-shot learning for novel categories with varying numbers of interaction data. \Cref{fig:adapt_change} shows the comparison with Dualafford  using the same number of budgets of interactions(a budget of 50 interactions in total).
Notably, our method's success rate at budget 0 is obtained via the semantic correspondence.
We can see that, even with a small number of interactions, our method consistently outperforms DualAfford, demonstrating superior initial performance obtained from the foundation model. And the success rate of our approach rises faster and higher as the interaction budget increases.
Particularly on the task \textbf{Closing}, 3 budgets of demonstrations lead to a dramatic increase (more than 30 \%), compared to the performance without finetuning.
These results highlight the advantage of our approach in achieving higher efficiency and better performance on unseen instances.

\subsection{Real-World Experiments}
\begin{figure}[t!]
    \includegraphics[width=0.95\linewidth]{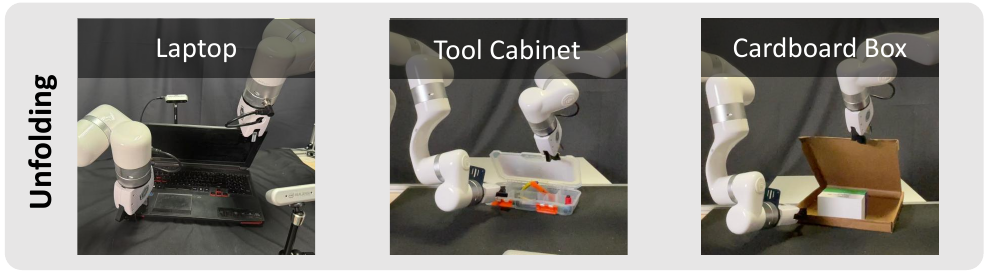}
    \centering
    \caption{Cross-category generalization for \textbf{unfolding} task.}
    \label{fig:generalization}
\end{figure} 
\Cref{fig:real_setting} shows our real-world setup and objects used. 
We conducted real-world experiments using two UFactory xArm6 equipped with a UFactory xArm Gripper, and a front-view RealSense D435 camera to capture RGB-D observations.
We use SAM~\cite{kirillov2023segment} to segment a mask of the target object from the scene and project the segmented image. We obtain the object's mesh using AR code~\cite{arcode2022} and estimate its pose using FoundationPose~\cite{wen2024foundationpose}. Subsequently, we sample the object's point cloud based on the estimated pose.
We first obtain the corresponding affordance on the 2D image, and then back-project the pixel-level points to get the 3D contact points based on depth information, as described in \Cref{sec:affordance_transfer}. Given the task, object point cloud, and the resulting points, our framework can propose bimanual action.
\Cref{fig:real_manip} shows our real experiments on different tasks. And \Cref{fig:generalization} illustrates our framework's superior generalization that it can successfully manipulate varied unseen instances across categories for the \textbf{unfolding} task.
\section{CONCLUSIONS}
We propose Bi-Adapt, a foundation-model-based framework for bimanual manipulation generalization with adaptation. Bi-Adapt transfers affordance across categories via semantic correspondence. We introduce a few-shot learning strategy to adapt the pre-trained model's action direction prediction on novel categories with limited interactions. Experiments evaluate our framework's improved performance on adaptation for unseen instances in novel categories.

\qheading{Limitation and Future Work.} Currently, this work is unable to complete more complex long-horizon tasks that involve multiple objects and combine short-horizon tasks. A unified framework could be considered for further research.







\bibliographystyle{IEEEtran}
\bibliography{IEEEabrv,refs}

\end{document}